\documentclass[journal,transmag]{IEEEtran}

\usepackage{times}
\usepackage{epsfig}
\usepackage{graphicx}
\usepackage{amsmath}
\usepackage{amssymb}

\usepackage{amsfonts}
\usepackage{algorithmic}
\usepackage{textcomp}
\usepackage{subfigure}
\usepackage{array}
\usepackage[table]{xcolor}
\usepackage{microtype}
\usepackage{nameref}
\usepackage{chngpage}
\usepackage{multirow}
\usepackage{booktabs}  
\usepackage{threeparttable} 
\ifCLASSINFOpdf
\else
\fi
\begin{document}
%
\title{Multi Point-Voxel Convolution (MPVConv) for Deep Learning on Point Clouds}


\author{\IEEEauthorblockN{Wei Zhou\IEEEauthorrefmark{1},
Xin Cao\IEEEauthorrefmark{1},
Xiaodan Zhang\IEEEauthorrefmark{1}, 
Xingxing Hao\IEEEauthorrefmark{1},
Dekui Wang\IEEEauthorrefmark{1}, and
Ying He\IEEEauthorrefmark{2}}
\IEEEauthorblockA{\IEEEauthorrefmark{1}School of Information Science and Technology,
Northwest University, Xi'an, China}
\IEEEauthorblockA{\IEEEauthorrefmark{2}School of Computer Science and Engineering, Nanyang Technological University, Singapore}
\thanks{Corresponding author: Wei Zhou (email: mczhouwei12@gmail.com), Xin Cao (email: xin\_cao@163.com), Xiaodan Zhang (email: xiaodanzhang@nwu.edu.cn).}}

\markboth{Journal of \LaTeX\ Class Files,~Vol.~14, No.~8, August~2015}%
{Shell \MakeLowercase{\textit{et al.}}: Bare Demo of IEEEtran.cls for IEEE Transactions on Cybernetics}
%



\IEEEtitleabstractindextext{%
\begin{abstract}

The existing 3D deep learning methods adopt either individual point-based features or local-neighboring voxel-based features, and demonstrate great potential for processing 3D data. However, the point based models are inefficient due to the unordered nature of point clouds and the voxel-based models suffer from large information loss. Motivated by the success of recent point-voxel representation, such as PVCNN, we propose a new convolutional neural network, called Multi Point-Voxel Convolution (MPVConv), for deep learning on point clouds. Integrating both the advantages of voxel and point-based methods, MPVConv can effectively increase the neighboring collection between point-based features and also promote independence among voxel-based features. Moreover, most of the existing approaches aim at solving one specific task, and only a few of them can handle a variety of tasks.
Simply replacing the corresponding convolution module with MPVConv, we show that MPVConv can fit in different backbones to solve a wide range of 3D tasks.  
Extensive experiments on benchmark datasets such as ShapeNet Part, S3DIS and KITTI for various tasks show that MPVConv improves the accuracy of the backbone (PointNet) by up to \textbf{36\%}, and achieves higher accuracy than the voxel-based model with up to \textbf{34}$\times$ speedups. In addition, MPVConv outperforms the state-of-the-art point-based models with up to \textbf{8}$\times$ speedups. Notably, our MPVConv achieves better accuracy than the newest point-voxel-based model PVCNN (a model more efficient than PointNet) with lower latency.
\end{abstract}

}

\maketitle

\IEEEdisplaynontitleabstractindextext

%
\IEEEpeerreviewmaketitle

\section{Introduction}

\IEEEPARstart{3}{D} deep learning for point clouds has received much attention in both industry and academia thanks to its potential for a wide range of applications, such as autonomous driving and robots.
The main technical challenges are due to the sparse and irregular nature of point clouds. 

The existing 3D deep learning methods can be roughly divided into voxel- and point-based methods according to the representations of point clouds.
The voxel-based methods convert the irregular and sparse point clouds into regular 3D grids so that the widely studied convolutional neural networks (CNN) can be applied directly~\cite{cciccek20163d,riegler2017octnet,zhou2018voxelnet}.
Since their performance heavily depends on the voxelization resolution, the voxel-based methods often suffer from large information loss when the resolution is low, as multiple adjacent points are quantized into the same grid, which are indistinguishable. Conversely, a high resolution volume would preserve the fine-detailed information, but requires significant amount of GPU memory and computation time due to the cubic complexity of volumes. 
In contrast, the point-based methods can handle high-resolution models, since they process the raw points in a local and separate manner~\cite{klokov2017escape,Li2019pointcnn,qi2017pointnet,qi2017pointnetplusplus,wang2019dynamic}. Taking advantage of the sparse representation of point clouds, the point-based methods consume much less GPU memory than the voxel-based methods. However, due to lack of regularity, they suffer from expensive random memory access and dynamic kernel computation during the point and its nearest neighbor searching~\cite{liu2019pvcnn}.

Motivated by the merits and limitations of each type of methods, several researchers proposed mixed representations to overcome the challenges of high accuracy demand and limited computational resources available on GPUs recently ~\cite{liu2019pvcnn,cherenkova2020pvdeconv,shi2021pv,zhang2020deep,tang2020searching,shi2020pv}. However, most of these point-voxel combined methods are only for solving a specific task with a point-voxel based framework. For example, PV-RCNN~\cite{shi2020pv} and PV-RCNN++~\cite{shi2021pv} focus on 3D object detection, and Pvdeconv~\cite{cherenkova2020pvdeconv} targets 3D auto-encoding CAD construction, FusionNet~\cite{zhang2020deep} is for semantic segmentation. Tang \emph{et al.} proposed a sparse point-voxel convolution for efficient 3D architecture searching~\cite{tang2020searching}. 
To our knowledge, there is no general point-voxel based method for solving different kind of tasks.

\begin{figure*}[t]
	\begin{center}
		\includegraphics[width=1\linewidth]{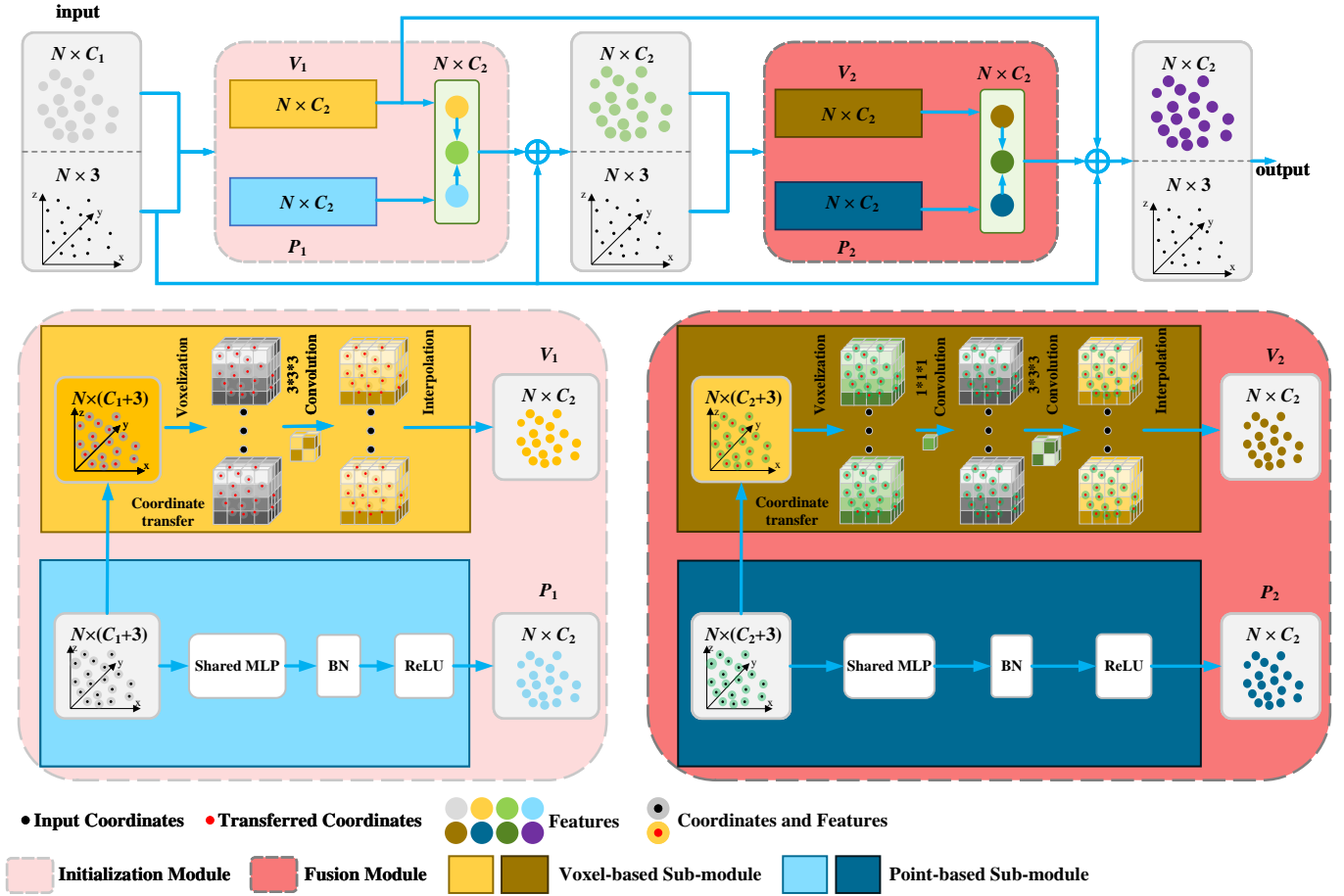}
	\end{center}
	\caption{MPVConv increases  neighborhood collection in point-based features and independence in voxel-based features by adopting 3D CNNs and MLPs on both types of features.}
	\label{fig:MPVconv_all}
\end{figure*}

To design effective deep neural networks for 3D analysis, one must take into consideration the performance, efficiency, as well as generality and flexibility for various tasks. In this paper, we propose Multi Point-Voxel Convolution (MPVConv) which takes the advantages of both the voxel- and point-based methods, and can work with different backbones for a wide range of 3D tasks. 
In the previous 3D CNN models, the point-based features are individual and the voxel-based features are based on local neighborhood. In contrast, our MPVConv conducts 3D CNN and MLP on both points and voxels by multi Point-Voxel modules. As a result, it increases not only the neighboring collection for point-based features, but also the independence among voxel-based features. Extensive experiments show that MPVCNN outperforms the state-of-the-art 3D CNN models in terms of accuracy and efficiency.

\section{Related Work}
\label{sec:related_work}

\textbf{Voxel-based 3D learning.} Inspired by the success of CNN on 2D images~\cite{tran2015learning,ren2016faster,redmon2016you}, researchers transfer point cloud representation to volumetric representation and attempt to adopt convolution over it~\cite{maturana2015voxnet,cciccek20163d,qi2016volumetric,le2018pointgrid,du2020vipnet}. As is known to all, the larger the voxel resolution is, the more detailed feature information is contained, and the calculation of 3D volumetric convolution increases exponentially. To slow down this problem, researchers adopt octree to construct efficient convolutional architectures to increase the efficiency of computation with high voxel resolutions~\cite{tatarchenko2017octree,riegler2017octnet,tatarchenko2017octree}.
State-of-the-art researches have demonstrated that volumetric representation can also be applied in 3D shape classification~\cite{wu20153d,wang2017cnn,le2018pointgrid}, 3D point cloud segmentation~\cite{graham20183d,meng2019vv,wang2019voxsegnet} and 3D object detection~\cite{zhou2018voxelnet}. Although the voxel-based methods have a great advantage in data structuring for 3D CNN, its computation efficiency is still greatly limited by the sizes of voxel resolution.

\textbf{Point-based 3D learning.} PointNet~\cite{qi2017pointnet}, the first deep neural network that directly consumes 3D points, adopts simple symmetric function (maxpooling) to achieve permutation invariance. Since PointNet does not take local geometry into consideration, one typical way to improve it is to adopt hierarchical aggregation for extracting local features~\cite{qi2017pointnetplusplus,klokov2017escape}. PointCNN~\cite{Li2019pointcnn}, SpiderCNN~\cite{xu2018spidercnn} and Geo-CNN~\cite{lan2019modeling} dynamically generate local geometric structures to capture points' neighboring features. RSNet~\cite{huang2018recurrent} adopts a lightweight local dependency module to efficiently model local structures of point clouds.
3D-GCN~\cite{lin2021learning}, DGCNN~\cite{wang2019dynamic}, Grid-GCN~\cite{xu2020grid}, SpecConv~\cite{wang2018local}, SPGraph~\cite{landrieu2018large}, GAC~\cite{wang2019graph} adopt graph convolutional networks to conduct 3D point cloud learning, while RS-CNN~\cite{liu2019relation}, PCNN~\cite{atzmon2018point}, SCN~\cite{xie2018attentional} and KCNet~\cite{shen2018mining} make use of geometric relations for point cloud analysis. Furthermore, SPLATNet~\cite{su2018splatnet} uses sparse bilateral convolutional layers to build the network, and SO-Net~\cite{li2018so} proposes permutation invariant architectures for learning with unordered point clouds. SSNet~\cite{thabet2020self} combines Morton-order curve and point-wise to conduct self-supervised learning. SPNet~\cite{liu2020self} uses a self-prediction for 3D instance and semantic segmentation of point clouds. RandLA-Net~\cite{hu2020randla} introduces a local feature aggregation module to preserve geometric details.
As pointed out in~\cite{liu2019pvcnn}, random memory access and dynamic kernel computation are the performance bottleneck of point-based methods.

\textbf{Point-voxel-based 3D learning.} Point-voxel-based methods combine the voxel- and point-based approaches. Aiming at 3D object detection, PV-RCNN~\cite{shi2020pv} attracts the multi-scaled 3D voxel CNN features by PointNet++~\cite{qi2017pointnetplusplus}, then the voxel- and point-based features are aggregated to a small set of representative points. Based on PV-RCNN, PV-RCNN++~\cite{shi2021pv} has improved the computation efficiency of point-based parts. Like PV-RCNN and PV-RCNN++ are specific to 3D object detection, other point-voxel-based 3D learning methods~\cite{cherenkova2020pvdeconv,zhang2020deep,tang2020searching} are specially designated only for 3D construction, 3D semantic segmentation or 3D structuring.
PVCNN~\cite{liu2019pvcnn} adopts interpolation to obtain the voxel-based CNN features, and then aggregates the point-based features and the interpolated voxel-based features as output. Comparing with the existing point-voxel-based methods~\cite{liu2019pvcnn,cherenkova2020pvdeconv,shi2021pv,zhang2020deep,tang2020searching,shi2020pv}, our method can deal with different kind of 3D tasks by applying MPVConv in various backbones with replacing the corresponding convolution module of backbones with MPVConv. 
Despite the resemblance of both combining voxel-based and point-based features, our work differs from~\cite{liu2019pvcnn}, which integrates both the advantages of the voxel- and point-based methods both into the voxel and point-based features, as we can effectively increase the neighboring collection between point-based features and also promote the independence among voxel-based features.
Moreover, we use shared MLPs to process the voxel-based features and adopt 3D CNN on the point-based features, thus improving the performances of various 3D deep learning tasks.

\section{Multi Point-Voxel Convolution}
\label{sec:MPVconv}


State-of-the-art 3D deep learning methods are based on either voxel-based CNN methods or point-based network. Generally speaking, 3D voxel-based CNN method owns good data locality and regularity for low-resolution point neighbors (voxel grids), while the point-based approaches are independent for each point, so it could capture high-resolution information. 

Inspired by point-voxel-based methods and convolution operation, we propose a new method, called Multi Point-Voxel Convolution (MPVConv), to conduct convolution on point clouds. Our method has two features that distinguishes itself from the existing point-voxel methods. First, the existing methods such as PV-RCNN~\cite{shi2020pv}, PV-RCNN++~\cite{shi2021pv}, Pvdeconv~\cite{cherenkova2020pvdeconv} and FusionNet~\cite{zhang2020deep} are designed for a specific task, while our method can be applied to different tasks by replacing the corresponding convolution in backbones with MPVConv. 
Second, the previous methods (e.g.~\cite{shi2020pv,shi2021pv,liu2019pvcnn}) apply 3D volumetric convolution to voxels and MLP operation to points separately, and then fuse neighboring voxel features and independent point features together as output. In contrast, MPVConv applies 3D CNN and MLP to both points and voxels by multi point-voxel modules,  yielding \textit{independent} voxel features and 
local-neighboring point features.

As illustrated in Figure~\ref{fig:MPVconv_all}, MPVConv consists of an initialization module (left) and a fusion module (right). 
Initialization module generates initial individual point-based features and initial local-neighboring voxel-based features. Fusion module attaches neighboring and independent attributes to the output point- and voxel-based features of initialization module respectively.
In each module, there are point-based and voxel-based sub-modules. 
The point-based sub-modules use shared MLPs to extract features for each point in an independent manner, hence they only consume a small amount of memory even for high-resolution point clouds. 
The voxel-based sub-modules aggregate features using volumetric CNNs and tri-linear interpolation to map voxel features back to points. Due to low voxel resolution adopted in MPVConv, the voxel sub-modules also have small memory consumption. 

\subsection{Initialization Module}
\label{sec:ini_neuron}

The initialization module generates initial individual point-based features and initial local-neighboring voxel-based features from the input 3D data $\textit{\textbf{X}} = \{\textit{\textbf{x}}_1,...,\textit{\textbf{x}}_n\} =\{(\textit{\textbf{p}}_1, \textit{\textbf{f}}_1),...,(\textit{\textbf{p}}_n, \textit{\textbf{f}}_n)\} \subseteq \mathbb{R}^{3+C_{1}}$, where $\textit{\textbf{p}}_i$ is the 3D point coordinates $\textit{\textbf{p}}_i = (x_i, y_i, z_i)$, it may also contain further information, e.g. RGB color and normal; $\textit{\textbf{f}}_i$ is the output feature of previous layer which includes $C_{1}$ channels. 

\subsubsection{Voxel-based Sub-module}
\label{sec:voxel_module}

3D CNN on voxel grid is a popular selection for state-of-the-art 3D deep learning researches. Due to its high regularity and efficient structuring with 3D CNN, we adopt the voxel-based sub-module to capture the initial neighboring information for initialization module. 

\noindent \textbf{Point Transformation.} Before we implement the voxel-based sub-module, we conduct point transformation to eliminate the influences from the translation and the scale variations of 3D point coordinates. Firstly, we calculate the mean point $\bar{\textit{\textbf{p}}}$ of the input 3D data, and translate each point with $\bar{\textit{\textbf{p}}}$ by $\textit{\textbf{p}}_{i} = \textit{\textbf{p}}_{i} - \bar{\textit{\textbf{p}}}$. Then we search the farthest point $\|\textit{\textbf{p}}\|_{max}$ as the radius, and transform the whole points into the unit sphere:
\begin{equation}
	\hat{\textit{\textbf{p}}}_{i} = \dfrac{\textit{\textbf{p}}_{i}}{2*\|\textit{\textbf{p}}\|_{max}} + 0.5,
\end{equation}
where $\hat{\textit{\textbf{p}}}_{i}$ is the transformed point coordinates. During the experiments, we find that feature transformation reduces the accuracy of our model. Therefore, we transform only point coordinates, and mark the transformed points as $\{\hat{\textit{\textbf{p}}}_{i}, \hat{\textit{\textbf{f}}}_{i}\}$.

\noindent \textbf{3D Voxel CNN.} After the point transformation, the coordinates range from 0 to 1. To match the voxel resolution, we scale the coordinates by a factor $\{r-1\}$, and mark the scaled coordinates as $\hat{\textit{\textbf{p}}}_{ri} (\hat{x}_{ri},\hat{y}_{ri},\hat{z}_{ri})$. Then the points $\hat{\textit{\textbf{p}}}_{ri}$ and its corresponding features $\hat{\textit{\textbf{f}}}_{i}$ are voxelized into the voxels with low spatial resolution of $r \times r \times r$, and the features of voxels are the mean features of all inside points:
\begin{equation}
	\textit{\textbf{F}}(u,v,w) = \frac{\sum_{i=1}^{n} \lfloor (\hat{x}_{ri},\hat{y}_{ri},\hat{z}_{ri}) \cdot \hat{\textit{\textbf{f}}}_{i}\rfloor}{n},
\end{equation}
where $\textit{\textbf{F}}(u,v,w)$ is the feature of voxel $(u,v,w)$, $n$ is the number of interior points inside the voxel, and $\lfloor\cdot\rfloor$ is the floor function.
We discuss the choice of voxelization resolution $r$ in Section~\ref{sec:ablation_study}.
Next, we adopt a series of $3 \times 3 \times 3$ 3D CNNs to aggregate the neighboring feature information of voxels. To capture the neighborhood information more accurately, we increase the feature channels to $C_2$ in 3D CNN implementation.

\noindent \textbf{Voxel Feature Interpolation.} Then we interpolate the voxel features into the common domain of point cloud, as we need to aggregate the information of voxel-based sub-module and point-based sub-module. The common operation for interpolation is the tri-linear interpolation and nearest-neighbor interpolation. Similar to the conduction in PVCNN, we adopt tri-linear interpolation to transform the voxel features into point features $\textit{\textbf{V}}_{1} = \{V_{11},...,V_{1n}\} \subseteq \mathbb{R}^{C_{2}}$.


\subsubsection{Point-based Sub-module}
\label{sec:point_module}

Although the voxel-based sub-module aggregates the neighboring feature information for the input 3D data, its extracted information is in a coarse manner with low voxel resolution. In order to make up for this defect, we adopt point-based shared MLPs (1D convolution with kernel of 1) to make efficient learning on each point for the input 3D data. For the sake of convenient aggregation of voxel-based sub-module and point-based sub-module, the output feature channels of point-based sub-module are consistent with the output channels of voxel-based sub-module. The output features are implemented with batch normalization~\cite{ioffe2015batch} and ReLU activation function~\cite{glorot2011deep}, and these point-based features are marked as $\textit{\textbf{P}}_{1} = \{P_{11},...,P_{1n}\} \subseteq \mathbb{R}^{C_{2}}$. 

When both the feature information of voxel-based and point-based sub-module are obtained, the initialization module will output $\textit{\textbf{V}}_{1}$ and $\textit{\textbf{P}}_{1}$ to the fusion module to strengthen the point-based and voxel-based features.

\subsection{Fusion Module}
\label{sec:trans_neuron}

The fusion module is used to strengthen the point- and voxel-based features from the initialization module. The enhancement of feature information is reflected in two aspects: (1) attaching the neighboring collection for the individual point-based features $\textit{\textbf{P}}_{1}$; (2) increasing the independence of the neighboring voxel-based features $\textit{\textbf{V}}_{1}$.

\noindent \textbf{Neighboring collection for individual point-based feature.} Fusion module obtains the input features from the output aggregation feature information of initialization module. In terms of network structures, the point-based sub-module in fusion module is the same as the one in initialization module, the voxel-based sub-module got a little differences. The voxel-based and point-based information ($\textit{\textbf{P}}_{1}, \textit{\textbf{V}}_{1}$) from the initialization module have aggregated together before inputting to the fusion module, but as we know, the voxel-based features $\textit{\textbf{V}}_{1}$ have carried out information exchanging between different channels by 3D CNN, and yet the point-based features $\textit{\textbf{P}}_{1}$ haven't exchanged. 
To enhance the neighboring collection for the point-based features $\textit{\textbf{P}}_{1}$ from the initialization module with 3D CNN in the voxel-based sub-module of the current fusion module, we firstly fuse $\textit{\textbf{P}}_{1}$ and $\textit{\textbf{V}}_{1}$ to obtain the fused features $\{\textit{\textbf{V}}_{1}+\textit{\textbf{P}}_{1}\}\subseteq \mathbb{R}^{C_{2}}$, then voxelize $\{\textit{\textbf{V}}_{1}+\textit{\textbf{P}}_{1}\}$ with low spatial resolution $r \times r \times r$. In order to further strengthen the neighboring collection for $\textit{\textbf{P}}_{1}$ and exchange the collection between different channels of $\textit{\textbf{V}}_{1}$, we set a 3D CNN with kernel of $1 \times 1 \times 1$ in the voxel-based sub-module of the current fusion module before the regular $3 \times 3 \times 3$ 3D CNN operations, thus we obtain the enhanced features $\textit{\textbf{V}}_{2} = \{V_{21},...,V_{2n}\} \subseteq \mathbb{R}^{C_{2}}$.

\noindent \textbf{Independence for the neighboring voxel-based features.} To attach the independent attribute for voxel-based features $\textit{\textbf{V}}_{1}$, and increase the fine granularity of point-based features $\textit{\textbf{P}}_{1}$ by the way, we adopt shared MLP to the fused features $\{\textit{\textbf{V}}_{1}+\textit{\textbf{P}}_{1}\}\subseteq \mathbb{R}^{C_{2}}$. The parameters and network structure are consistent with the point-based sub-module of initialization module, and here we obtain the enhanced individual features $\textit{\textbf{P}}_{2} = \{P_{21},...,P_{2n}\} \subseteq \mathbb{R}^{C_{2}}$.
After the completion of point-based sub-module and voxel-based sub-module in the fusion module, next is to output the aggregating information. 
To output the final features, in addition to the fusion module's aggregation information, the initialization module's voxel-based features are also aggregated to the output, and thus we output $\{\textit{\textbf{V}}_{1}+\textit{\textbf{V}}_{2}+\textit{\textbf{P}}_{2}\}\subseteq \mathbb{R}^{C_{2}}$ as the final features. See  Section~\ref{sec:ablation_study} for the discussion on output aggregating features. 

\section{Experimental Results}
\label{sec:exp}

This section presents the implementation details and compares our method with the state-of-the-art frameworks on benchmark datasets for various tasks, including ShapeNet Parts (object part segmentation)~\cite{chang2015shapenet}, S3DIS (indoor scene segmentation)~\cite{armeni20163d,armeni2017joint} and KITTI (3D object detection)~\cite{geiger2013vision}. We also conduct ablation study to study the performance of MPVConv.

\subsection{Implementation Details}
\label{sec:MPV_detail}

\textbf{Initialization module.} We implemented MPVConv in PyTorch. In the initialization module, the voxel-based sub-module contains two $3 \times 3 \times 3$ 3D CNNs with stride 1 and padding 1. Each 3D CNN is followed by 3D batch normalization~\cite{ioffe2015batch} and the leaky ReLU activation function~\cite{maas2013rectifier}. The point-based sub-module contains shared MLPs which is a 1D CNN with kernel of 1 which converts the feature channels to make it consistent with the output feature channels of voxel-based sub-module, next to the 1D CNN is 1D batch normalization~\cite{ioffe2015batch} and ReLU activation function~\cite{glorot2011deep}.

\textbf{Fusion module.} The fusion module follows the initialization module and takes its aggregation-fused feature as input. Its voxel-based sub-modules contain one $1 \times 1 \times 1$ 3D CNN with stride 1 and padding 0, and the rest part is the same as the one in the voxel-based sub-modules of the initialization module. The point-based sub-module also keeps the output feature channels consistent with its voxel-based sub-module.

\subsection{Part Segmentation}
\label{sec:shapenet}

\textbf{Dataset.} We apply  MPVConv to 3D part segmentation on ShapeNet Part~\cite{chang2015shapenet}, which consists of 16,881 3D shapes in 16 categories. We sample 2048 points from each shape as the training data. To make fair comparisons with others, we follow the same evaluation scheme as adopted in PVCNN~\cite{liu2019pvcnn}, PointCNN~\cite{Li2019pointcnn}, and SSCN~\cite{graham20183d} in our experiments. 

\textbf{Architecture.} Using PointNet~\cite{qi2017pointnet} as the backbone, we build MPVCNN by replacing its shared MLP layers with the proposed MPVConv layers. 

\textbf{Training.} We use the cross-entropy loss function, set the batch size to 8, and adopt the ADAM optimizer~\cite{kingma2014adam} with learning rate of 0.001 for 200 epochs. We carried out the training process on a RTX 2080Ti GPU.

\begin{figure*}[t]
	\begin{center}
		\includegraphics[width=0.95\linewidth]{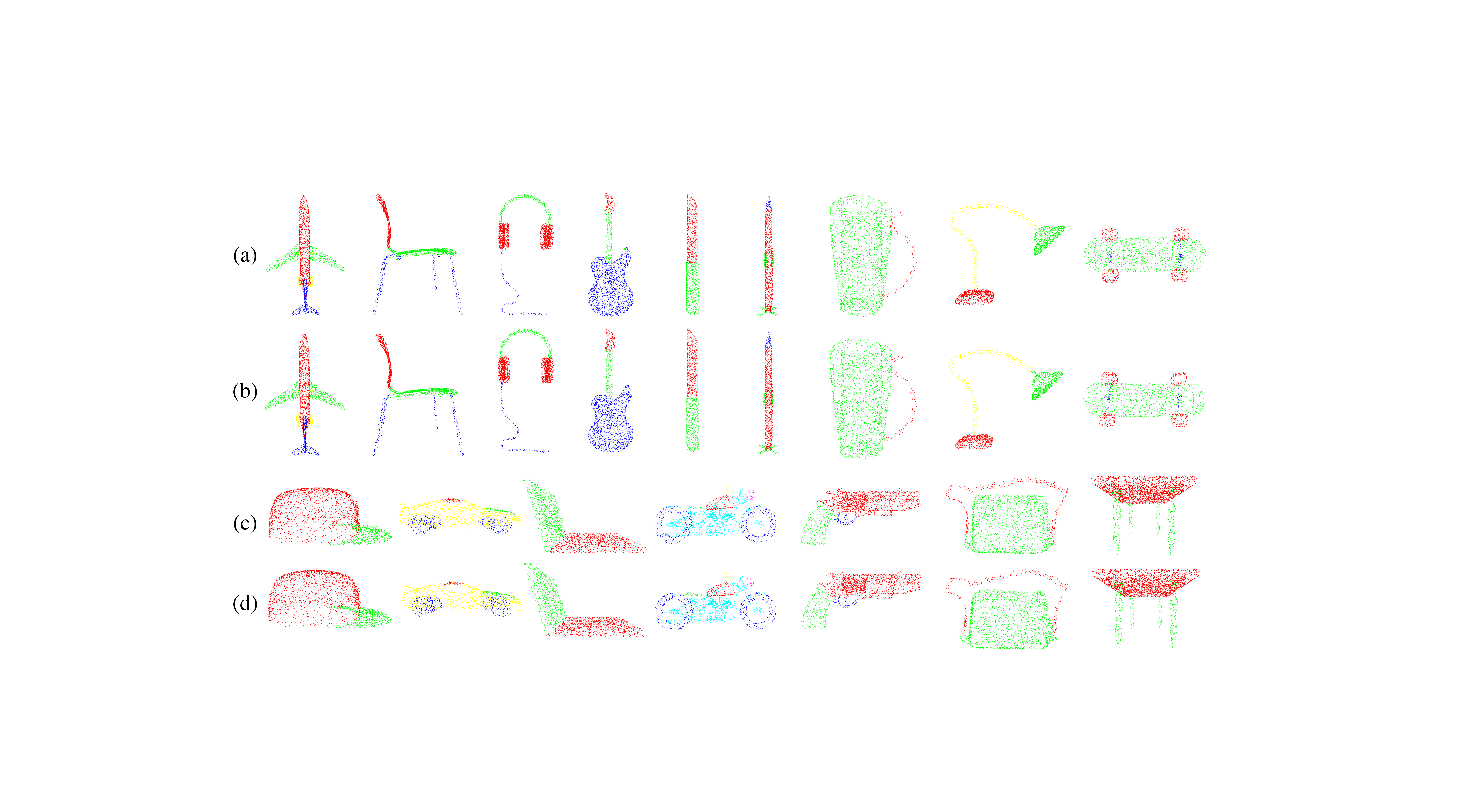}
	\end{center}
	\caption{Segmentation results on ShapeNet Part. (a)(c) Ground truth. (b)(d) Our results.}
	\label{fig:vis_shapenet}
\end{figure*}

\textbf{Results.} We use the mean intersection-over-union (mIoU) as the evaluation metric. Same as the evaluation scheme in PointNet, we calculate the IoU of each shape by averaging the same shape parts' IoUs of the 2874 test models. We then compute the global mIoU by averaging the IoU of all shapes. We compare our model against the state-of-the-art point-based methods~\cite{klokov2017escape,qi2017pointnet,li2018so,xie2018attentional,su2018splatnet,shen2018mining,huang2018recurrent,qi2017pointnetplusplus,wang2019dynamic,atzmon2018point,xu2018spidercnn,xu2020geometry,lin2021learning,Li2019pointcnn,thomas2019kpconv,liu2020self,wen2020cf,liu2019relation,liu2020fg}, voxel-based methods~\cite{cciccek20163d} and the recent point-voxel-based model~\cite{liu2019pvcnn}.
To better balance time efficiency and accuracy, we also reduce the output feature channels to 50\% and 25\%, and call the downsized networks  $\textbf{MPVCNN}_{(0.5 \times Ch)}$ and $\textbf{MPVCNN}_{(0.25 \times Ch)}$ respectively. Table~\ref{tab:shapenet} and Figure~\ref{fig:vis_shapenet} show the results of MPVCNN on the ShapeNet part dataset. See also the supplementary material for more visual results.
MPVCNN outperforms the PointNet backbone with 3.3\% increase of mIoU, even if we reduce the latency by 8.8\%, we can still obtain 2.2\% mIoU ahead of PointNet. Comparing with the voxel-based 3D-UNet and point-based SpiderCNN, our quarter version is $\textbf{34} \times$ and $\textbf{8} \times$ faster respectively, while achieving the better accuracy. In addition, MPVConv also outperforms other SOTA point-based methods (3D-GCN) with better accuracy by a large margin of 1.4\%.
Notably, even compared with the voxel-point-based PVCNN, we still gain higher mIoU by 0.3\% with a small amount of time efficiency lost, and PVCNN is a state-of-the-art method aimed at speed improvement.

\renewcommand{\arraystretch}{1.0}  
\begin{table}[ht]
	\centering  
	\fontsize{9}{10}\selectfont  
	\setlength\tabcolsep{1pt}
	\begin{threeparttable}  
		\caption{Evaluation results of part segmentation on ShapeNet Part dataset. P, V and PV stand for point-, voxel- and point-voxel-based.
		}  
		\label{tab:shapenet}  
		\begin{tabular}{lccccccccccccccccccccccc}  
			\toprule  
			Method&Type&Batch&Input &mIoU&GPU&Latency\cr  
			      &    &Size&Size&(\%) &(GB) & (ms)\cr  
			\midrule
			Kd-Net~\cite{klokov2017escape}&P&8& 4K&82.3&-&-\cr
			PointNet~\cite{qi2017pointnet}&P&8&2K&83.7&\textbf{1.5}&21.7\cr
			3D-UNet~\cite{cciccek20163d}&V&8&$96^3$&84.6&8.8&682.1\cr 	
			SO-Net~\cite{li2018so}&P&8&1K&84.6&-&-\cr
			SCN~\cite{xie2018attentional}&P&8&1K&84.6&-&-\cr
			SPLATNet~\cite{su2018splatnet}&P&-&-&84.6&-&-\cr
			KCNet~\cite{shen2018mining}&P&8&2K&84.7&-&-\cr
			RSNet~\cite{huang2018recurrent}&P&8&2K&84.9&0.8&74.6\cr
			PointNet++~\cite{qi2017pointnetplusplus}&P&8&2K&85.1&2.0&77.9\cr
			DGCNN~\cite{wang2019dynamic}&P&8&2K&85.1&2.4&87.8\cr
			PCNN~\cite{atzmon2018point}&P&8&2K&85.1&-&-\cr
			SpiderCNN~\cite{xu2018spidercnn}&P&8&2K&85.3&6.5&170.7\cr
			GSNet~\cite{xu2020geometry}&P&-&-&85.3&-&-\cr
			3D-GCN~\cite{lin2021learning}&P&8&2K&85.3&-&-\cr
			$\textbf{MPVCNN}_{(0.25 \times Ch)}$&PV&8&2K&85.5&1.7&\textbf{19.8}\cr
			$\textbf{MPVCNN}_{(0.5 \times Ch)}$&PV&8&2K&85.7&2.1&31.0\cr
			PointCNN~\cite{Li2019pointcnn}&P&8&2K&86.1&2.5&135.8\cr
			SPNet~\cite{liu2020self}&P&8&2K&86.2&-&-\cr
			CF-SIS~\cite{wen2020cf}&P&8&2K&86.2&-&-\cr
			PVCNN~\cite{liu2019pvcnn}&PV&8&2K&86.2&1.6&50.7\cr
			RS-CNN~\cite{liu2019relation}&P&8&2K&-&-&-\cr
			KPConv~\cite{thomas2019kpconv}&P&8&2K&86.4&-&-\cr
			$\textbf{MPVCNN}_{(1 \times Ch)}$&PV&8&2K&\textbf{86.5}&2.8&81.9\cr
			\bottomrule  
		\end{tabular}  
	\end{threeparttable} 
\end{table}

\subsection{Indoor Scene Segmentation}
\label{sec:s3dis}

\textbf{Dataset.} Stanford large-scale 3D Indoor Spaces (S3DIS) Dataset~\cite{armeni20163d,armeni2017joint} is a benchmark used for semantic indoor scene segmentation. S3DIS consists of 6 3D scanning indoor areas which totally includes 272 rooms. Similar to \cite{liu2019pvcnn,Li2019pointcnn,tchapmi2017segcloud}, we take Area-1, 2, 3, 4, 6 as the training set, and Area-5 (which is the only one that is not overlapping with others) as the testing set. We follow the data preprocessing and the evaluation criteria as in~\cite{Li2019pointcnn} before training and sample each block with 4096 points for training. 

\textbf{Architecture.} In this task, we also use PointNet as the backbone for MPVCNN, and PointNet++ for MPVCNN++. Similar to the part segmentation  experiments in Section~\ref{sec:shapenet}, we also design a compact version by reducing the output feature channels to 12.5\%, 25\% and 50\% in MPVCNN, and 50\% in MPVCNN++.

\textbf{Training.} The batch size, optimizer, learning rate and criterion are set the same as what we have done in ShapeNet part experiments. The number of epoch is set to 50. It is also implemented on a single RTX 2080Ti GPU.

\begin{figure*}[t]
	\begin{center}
		\includegraphics[width=0.95\linewidth]{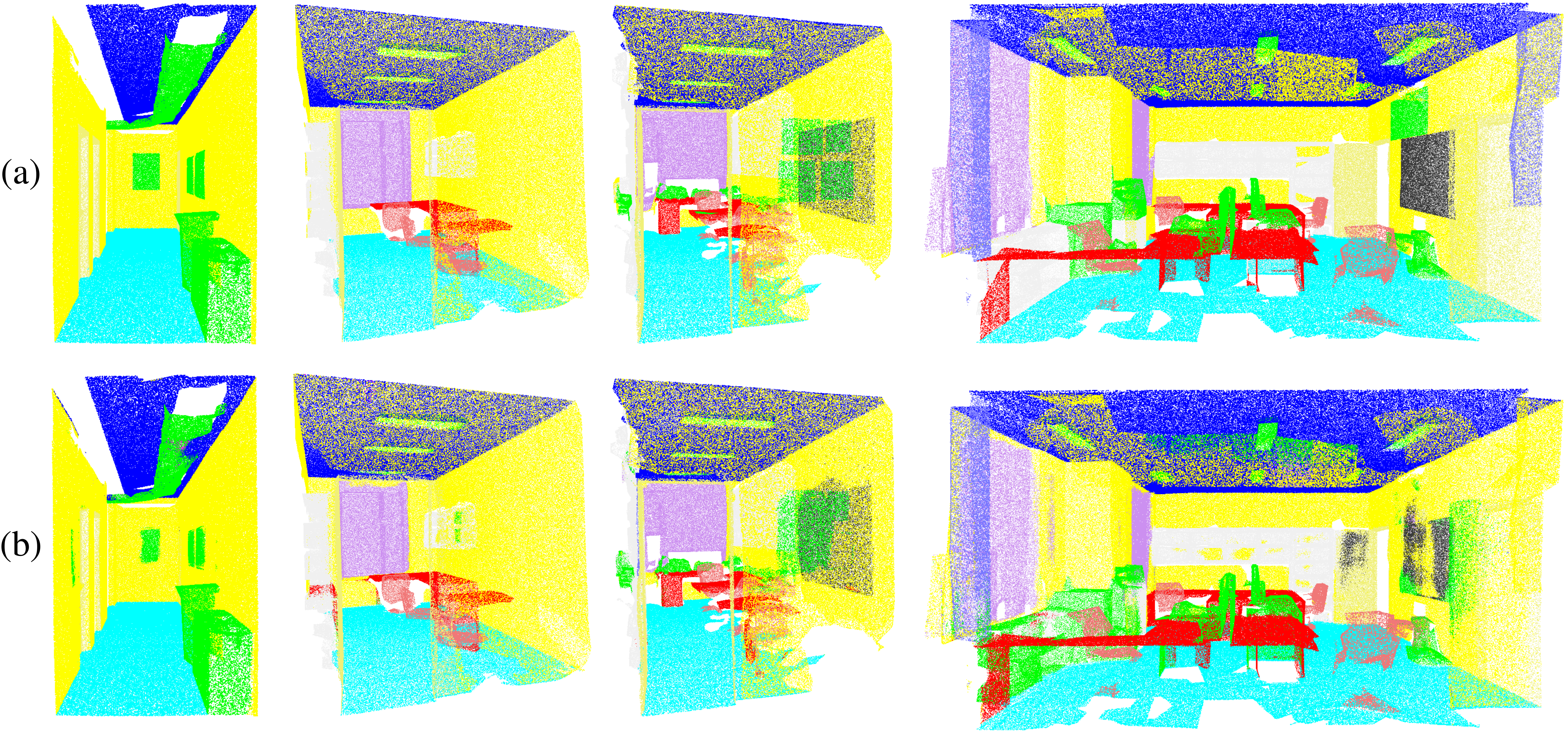}
	\end{center}
	\caption{Segmentation results on S3DIS Area-5. (a) Ground truth. (b) Our results.}
	\label{fig:vis_s3dis}
\end{figure*}

\textbf{Results.} Apart from mIoU, mean accuracy (mAcc) is also used to evaluate the performances of our proposed model. In addition to comparing with the state-of-the-art point-based~\cite{qi2017pointnet,wang2019dynamic,lin2021learning,huang2018recurrent,qi2017pointnetplusplus,tatarchenko2018tangent,zhao2020jsnet,thabet2020self,Li2019pointcnn,xu2020grid,liu2020self,wen2020cf} and voxel-based methods~\cite{cciccek20163d}, we also compare with the newest point-voxel-based model~\cite{liu2019pvcnn}. 
Table~\ref{tab:s3dis} shows the results of all methods on S3DIS dataset, and Figure~\ref{fig:vis_s3dis} presents the visualization results, more visualization results of S3DIS are presented in the supplementary material.
Specifically, the batch size of KPConv~\cite{thomas2019kpconv} is set to 1 since KPConv takes up too much memory to run.
Our MPVCNN improves the mIoU of the backbone (PointNet) by more than \textbf{36\%}, and MPVCNN++ outperforms its backbone (PointNet++) by a large margin in mIoU of more than \textbf{17\%}. Notably, The compact MPVCNN of 25\% feature channel outperforms the voxel-based 3D-UNet in accuracy with more than \textbf{11}$\times$ lower latency, and the point-based DGCNN by more than \textbf{15\%} in mIoU with $\textbf{3} \times$ lower latency. Moreover, the full model of MPVCNN outperforms the state-of-the-art point-based model (Grid-GCN and PointCNN) with 2.4\% increasing of the mIoU, and we increase the speed by nearly \textbf{4}$\times$ compared with PointCNN. At the same time, MPVCNN also outperforms the newest voxel-point-based method (PVCNN) both in mIoU by 4.5\% and mAcc by 1.3\%. Remarkably, the compact version of MPVCNN++ is faster than the extremely efficient voxel-point-based PVCNN++, and also outperforms it in accuracy by a large margin of 2\%.The full MPVCNN++ can improve the performance more than \textbf{4}\% in mIoU compared with PVCNN++. Overall, our MPVConv owns efficiency, small GPU memory requirements and good extensibility without sacrificing the performance of accuracy.

\renewcommand{\arraystretch}{1.0}  
\renewcommand\tabcolsep{3.0pt}
\begin{table}[ht]
	\centering  
	\fontsize{8}{9}\selectfont  
	\begin{threeparttable}  
		\caption{Evaluation results of indoor scene segmentation on S3DIS dataset.}  
		\label{tab:s3dis}  
		\begin{tabular}{lccccccccccccccccccccccc}  
			\toprule  
			Method&Type&Batch&Input&mIoU&mAcc&GPU&Latency\cr 
			&&Size&Size&(\%)&(\%)&(GB)&(ms)\cr 
			\midrule
			PointNet~\cite{qi2017pointnet}&P&8&4K&42.97&82.54&\textbf{0.6}&\textbf{20.9}\cr
			DGCNN~\cite{wang2019dynamic}&P&8&4K&47.94&83.64&2.4&178.1\cr
			KPConv~\cite{thomas2019kpconv}&P&1&4K&48.58&-&10.7&-\cr
			3D-GCN~\cite{lin2021learning}&P&8&4K&51.90&84.60&-&-\cr
			RSNet~\cite{huang2018recurrent}&P&8&4K&51.93&-&1.1&111.5\cr
			PointNet++~\cite{qi2017pointnetplusplus}&P&8&4K&52.28&-&-&-\cr
			TanConv~\cite{tatarchenko2018tangent}&P&8&4K&52.8&85.5&-&-\cr
			3D-UNet~\cite{cciccek20163d}&V&8&$96^3$&54.93&86.12&6.8&574.7\cr
			JSNet~\cite{zhao2020jsnet}&P&8&4K&54.5&87.7&-&-\cr
			SSNet~\cite{thabet2020self}&P&-&-&55.00&61.20&-&-\cr
			$\textbf{MPVCNN}_{(0.25 \times Ch)}$&PV&8&4K&55.30&86.91&2.3&51.9\cr 
			PVCNN~\cite{liu2019pvcnn}&PV&8&4K&56.12&86.66&1.3&47.3\cr
			PointCNN~\cite{Li2019pointcnn}&P&16&2K&57.26&85.91&4.6&282.3\cr
			Grid-GCN~\cite{xu2020grid}&P&8& 4K&57.75&86.94&-&25.9\cr
			$\textbf{MPVCNN}_{(1 \times Ch)}$&PV&8& 4K&58.63&87.75&4.3&72.7\cr
			SPNet~\cite{liu2020self}&P&8&4K&58.80&65.90&-&-\cr
			CF-SIS~\cite{wen2020cf}&P&8&4K&58.90&67.30&-&-\cr
			PVCNN++~\cite{liu2019pvcnn}&PV&4&8K&58.98&87.12&0.8&69.5\cr
			$\textbf{MPVCNN++}_{(0.5 \times Ch)}$&PV&4& 8K&60.17&88.76&2.7&52.6\cr
			$\textbf{MPVCNN++}_{(1 \times Ch)}$&PV&4&8K&\textbf{61.51}&\textbf{89.31}&4.3&72.7\cr
			\bottomrule  
		\end{tabular}  
	\end{threeparttable} 
\end{table}

\subsection{3D Detection}
\label{sec:kitti}

\renewcommand{\arraystretch}{1.2}  
\begin{table*}[ht]  
	\centering  
	\fontsize{9}{10}\selectfont  
	\begin{threeparttable}  
		\caption{KITTI}  
		\label{tab:kitti}  
		\begin{tabular}{lcccccccccccc}  
			\toprule  
			\multicolumn{1}{c}{}&  
			\multicolumn{3}{c}{Car}&\multicolumn{3}{c}{Pedestrian}&\multicolumn{3}{c}{Cyclist}&\multicolumn{2}{c}{Efficiency}\cr  
			\cmidrule(lr){2-4} \cmidrule(lr){5-7}  \cmidrule(lr){8-10} \cmidrule(lr){11-12}
			&Easy &Mod.&Hard&Easy&Mod.&Hard&Easy&Mod.&Hard&GPU Mem.&Latency\cr  
			\midrule  
			F-PointNet~\cite{qi2018frustum}&85.24&71.63&63.79&66.44&56.90&50.43&77.14&56.46&52.79&1.3GB&29.1ms\cr  
			F-PointNet++~\cite{qi2018frustum}&84.72&71.99&64.20&68.40&60.03&52.61&75.56&56.74&53.33&2.0GB&105.2ms\cr  
			F-PVCNN~\cite{liu2019pvcnn}&85.25&72.12&64.24&70.60&61.24&56.25&78.10&57.45&53.65&1.4GB&58.9ms\cr  
			F-MPVCNN&\textbf{85.66}&\textbf{72.63}&\textbf{64.62}&\textbf{71.12}&\textbf{62.34}&\textbf{57.13}&\textbf{79.85}&\textbf{58.28}&\textbf{54.62}&2.2GB&100.0ms\cr
			\bottomrule  
		\end{tabular}  
	\end{threeparttable}  
\end{table*}

\textbf{Dataset.} We evaluate the performance of our model for 3D detection on KITTI~\cite{geiger2013vision}, a benchmark for autonomous driving which contains 7,481 training and 7,518 test samples. To make fair comparison, we follow the experimental setup as Qi \emph{et al.}~\cite{qi2018frustum}. We split the training samples to the \textit{train} set with 3,712 samples and the \textit{val} set with 3,769 samples.

\textbf{Architecture.} Without changing the whole network, we set F-PointNet~\cite{qi2018frustum} as the backbone by replacing the shared MLP layers of the instance segmentation network with our MPVConv to generate F-MPVCNN. 

\textbf{Training.} The same setting for training as in part segmentation and indoor scene segmentation, except that epoch is set to 209 and batch size is set to 32.

\textbf{Results.} Mean average precision (mAP) is adopted to evaluate our model. We compare our model against F-PointNet, F-PointNet++ and F-PVCNN (whose backbone is also F-PointNet). Table~\ref{tab:kitti} shows the experimental results on KITTI. Our F-MPVCNN outperforms all methods in all classes, and ours improves the mAP of F-PointNet (backbone) by up to \textbf{13.2\%}. Compared with F-PointNet++, we improve the accuracy of it by a large margin up to 8.6\% in the hard pedestrian class, and we are faster than it.

\subsection{Ablation Study}
\label{sec:ablation_study}

We conducted ablation experiments to understand the contributions of components to MPVConv. We trained the network and tested it on the ShapeNet part datasets~\cite{chang2015shapenet}, and considered half of the feature channels to output as it is a  trade-off between accuracy and latency.  

\textbf{Fusion module.} We tried to enlarge the voxel resolution and directly deepen 3D CNN network of voxel-based sub-module in initialization module before we think about constructing the fusion module for the whole MPVConv.
As presented in Table~\ref{tab:cnn_res} (row 2), increasing the voxel resolution can slightly improve the performance of network in segmentation task, but at a high cost of GPU memory consumption and latency. Conversely, Table~\ref{tab:cnn_res} (row 3) shows that the accuracy of deepening network structure by 3D convolution is reduced on the contrary, and the GPU memory and latency increased also, but the extra computational cost can be neglected.
So we tried another way to deepen the network, that is the fusion module. As shown in Table~\ref{tab:cnn_res} (row 4), the fusion module improves mIoU by 0.3\% for MPVConv without consuming much more GPU memory and latency. 

\renewcommand{\arraystretch}{1.0}  
\renewcommand\tabcolsep{3.0pt}
\begin{table}[ht]
	\centering  
	\fontsize{9}{11}\selectfont  
	\begin{threeparttable}  
		\caption{Effects of the fusion module.}  
		\label{tab:cnn_res}  
		\begin{tabular}{lcccccccccccccccccccccc}  
			\toprule  
			Model&mIoU&GPU Mem.&Latency\cr  
			\midrule
			Init. module&85.50&1.97GB&22.3ms\cr
			Init. module (1.5$\times$R)&85.55&2.35GB&37.1ms\cr
			Init. module (3$\times$Conv3D)&85.33&2.08GB&25.1ms\cr
			Init. + Fusion modules&\textbf{85.76}&2.11GB&31.0ms\cr
			\bottomrule  
		\end{tabular}  
	\end{threeparttable} 
\end{table}  

\textbf{Features.} Table~\ref{tab:feature_selection} reveals the effects of various combinations of the four feature components of MPVConv. Model A, with only feature $\textbf{\textit{V}}_{2}$, has performance downgrade, as the voxel-based feature $\textbf{\textit{V}}_{2}$ is not enough for 3D learning. The feature aggregations of D, F, G and H improve the performances of MPVConv, and the combination of $\textbf{\textit{V}}_{1}+\textbf{\textit{V}}_{2}+\textbf{\textit{P}}_{2}$ achieves the best performance.

\renewcommand{\arraystretch}{1.0}  
\renewcommand\tabcolsep{3.0pt}
\begin{table}[ht]
	\centering  
	\fontsize{9}{11}\selectfont  
	\begin{threeparttable}  
		\caption{Effects of different features for MPVConv.}  
		\label{tab:feature_selection}  
		\begin{tabular}{cccccccccccccccccccccccc}  
			\toprule  
			Model\quad&\quad$\textbf{\textit{V}}_{1}$\quad&$\quad\textbf{\textit{P}}_{1}$\quad&\quad$\textbf{\textit{V}}_{2}$\quad&\quad$\textbf{\textit{P}}_{2}$\quad&\quad mIoU\cr  
			\midrule
			A\quad&\quad \quad&\quad \quad&\quad \checkmark \quad&\quad \quad&\quad 85.39\cr
			B\quad&\quad \checkmark \quad&\quad \checkmark \quad&\quad \quad&\quad \quad&\quad 85.50\cr
			C\quad&\quad \quad&\quad \checkmark \quad&\quad \checkmark \quad&\quad \quad&\quad 85.43\cr
			D\quad&\quad \quad&\quad \quad&\quad \checkmark \quad&\quad \checkmark \quad&\quad 85.58\cr
			E\quad&\quad \checkmark \quad&\quad \checkmark \quad&\quad \checkmark \quad&\quad \quad&\quad85.46\cr
			F\quad&\quad \quad&\quad \checkmark \quad&\quad \checkmark \quad&\quad \checkmark \quad&\quad85.54\cr
			G\quad&\quad \checkmark \quad&\quad \quad&\quad \checkmark \quad&\quad \checkmark \quad&\quad \textbf{85.76}\cr
			H\quad&\quad \checkmark \quad&\quad \checkmark \quad&\quad \checkmark \quad&\quad \checkmark \quad&\quad85.57\cr	
			\bottomrule  
		\end{tabular}  
	\end{threeparttable} 
\end{table}

\textbf{1}$\times$\textbf{1}$\times$\textbf{1 3D CNN.} As  Table~\ref{tab:1conv} shows, 1$\times$1$\times$1 3D CNN can strengthen the information association and nonlinearity between different feature channels of point-based and voxel-based features, thus can improve the 3D learning ability of the network. With almost no increasing of latency and GPU memory, 1$\times$1$\times$1 3D CNN helps MPVConv to achieve better accuracy. 

\renewcommand{\arraystretch}{1.0}  
\renewcommand\tabcolsep{3.0pt}
\begin{table}[ht]
	\centering  
	\fontsize{9}{11}\selectfont  
	\begin{threeparttable}  
		\caption{Effects of 1$\times$1$\times$1 3D CNN for MPVConv.}  
		\label{tab:1conv}  
		\begin{tabular}{lcccccccccccccccccccccc}  
			\toprule  
			Model&mIoU&GPU Mem.&Latency\cr  
			\midrule
			MPVConv&85.72&2.024GB&31.2ms\cr
			MPVConv (1$\times$1$\times$1 CNN)&\textbf{85.76}&2.033GB&31.6ms\cr
			\bottomrule  
		\end{tabular}  
	\end{threeparttable} 
\end{table}  



\section{Conclusion}
\label{sec:conclusion}

We presented MPVConv, a 3D convolution neural network, for deep learning on point clouds. Combining both points and voxels, our method  increases the neighboring collection between point-based features and promote the independence among voxel-based features with efficient convolutions. Experimental results on several benchmark datasets show that our method can improve the performance of common 3D tasks, such as segmentation and detection.


\bibliographystyle{IEEEtran}
\bibliography{egbib}

\ifCLASSOPTIONcaptionsoff
  \newpage
\fi

\begin{IEEEbiography}[{\includegraphics[width=1in,height=1.25in,clip,keepaspectratio]{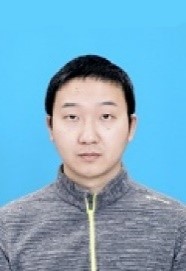}}]{Wei Zhou}
received his B.E. degrees in Electronic and Information Engineering from Xidian University in 2013. From 2013 to 2018 he is doing 5-year-system Ph.D. candidate in Xi’an Institute of Optics and Precision Mechanics of CAS \& University of Chinese Academy of Sciences. From 2016 to 2017, he was visiting Fraunhofer IGD \& TU Darmstadt. Since 2019, he has been a lecturer with School of Information Science and Technology, Northwest University, Xi'an, China. His main research interests include image restoration, 3D semantic segmentation, 3D recognition, 3D feature matching and other point-cloud-based and image-based researches.
\end{IEEEbiography}

\begin{IEEEbiography}[{\includegraphics[width=1in,height=1.25in,clip,keepaspectratio]{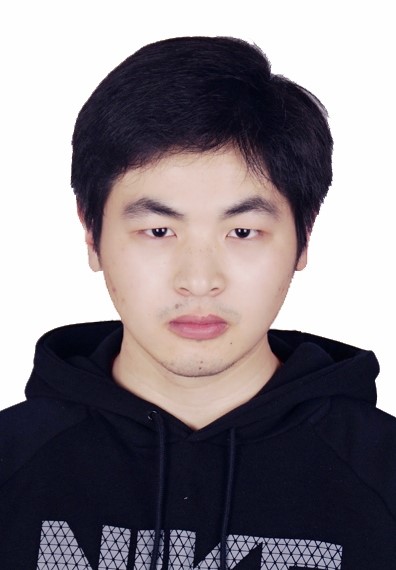}}]{Xin Cao} received the B.S. degrees in electronic engineering from Xidian University, Xi’an, China, in 2011 and the Ph.D. degree in pattern recognition and intelligent system from Xidian University, Xi’an, China, in 2016. 
Since 2016, he has been a lecturer with School of Information Science and Technology, Northwest University, Xi’an, China. His research interests include medical image analysis and  optical molecular imaging. 
\end{IEEEbiography}

\begin{IEEEbiography}[{\includegraphics[width=1in,height=1.25in,clip,keepaspectratio]{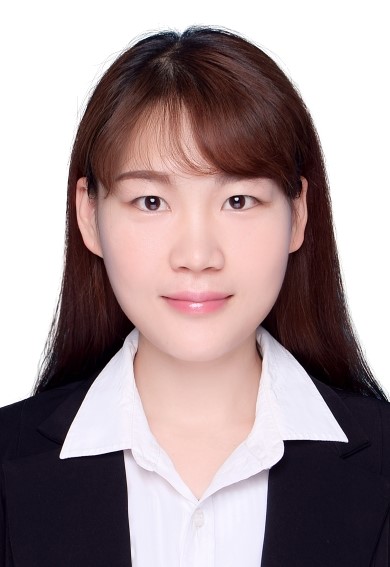}}]{Xiaodan Zhang}
received the B.Eng. degree in electronic information engineering and Ph.D. degree in intelligence information processing from Xidian University, Xi’an, China, in 2014 and 2019. She is currently a faculty member with Northwest University, Xi’an, China. Her current research interests include image aesthetic quality assessment, visual attention, and computationally modeling of human visual system.
\end{IEEEbiography}

\begin{IEEEbiography}[{\includegraphics[width=1in,height=1.25in,clip,keepaspectratio]{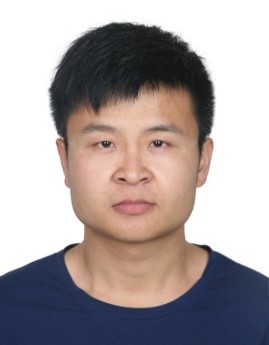}}]{Xingxing Hao} received the B.S. degree in intelligent science and technology from Xidian University, Xi'an, China in 2014. Now, he is pursuing the Ph.D. degree in circuits and systems from the School of Artificial Intelligence, Xidian University. His research interests include image processing, combinatorial optimization and evolutionary algorithms. 
\end{IEEEbiography}

\begin{IEEEbiography}[{\includegraphics[width=1in,height=1.25in,clip,keepaspectratio]{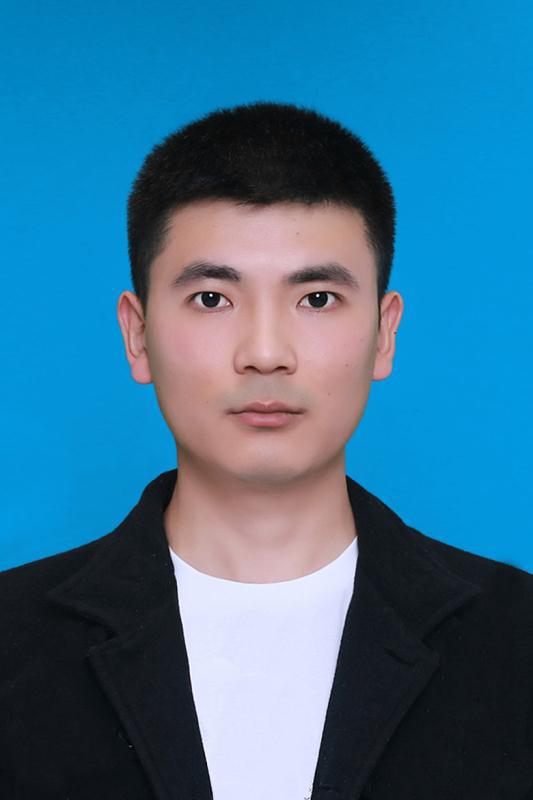}}]{Dekui Wang}
received the Ph.D. and B.S. degree in computer science and technology from Xidian University, Xi’an, China. He is working at School of Information Science and Technology, Northwest University, Xi’an, China. His current research interests include very large-scale integrated physical design automation with emphasis on synthesis, placement, routing, timing analysis and image processing.
\end{IEEEbiography}

\begin{IEEEbiography}[{\includegraphics[width=1in,height=1.25in,clip,keepaspectratio]{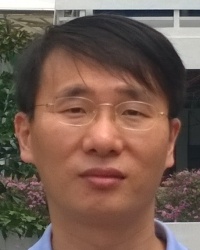}}]{Ying He}
received the BS and MS degrees in electrical engineering from Tsinghua University, China, and the PhD degree in computer science from Stony Brook University. He is currently an Associate Professor with School of Computer Science and Engineering, Nanyang Technological University, Singapore. His research interests fall into the general areas of visual computing and he is particularly interested in the problems which require geometric analysis and computation.
\end{IEEEbiography}




\clearpage
\appendices
\section{}
\begin{figure*}[t]
	\begin{center}
		\includegraphics[width=0.9\linewidth]{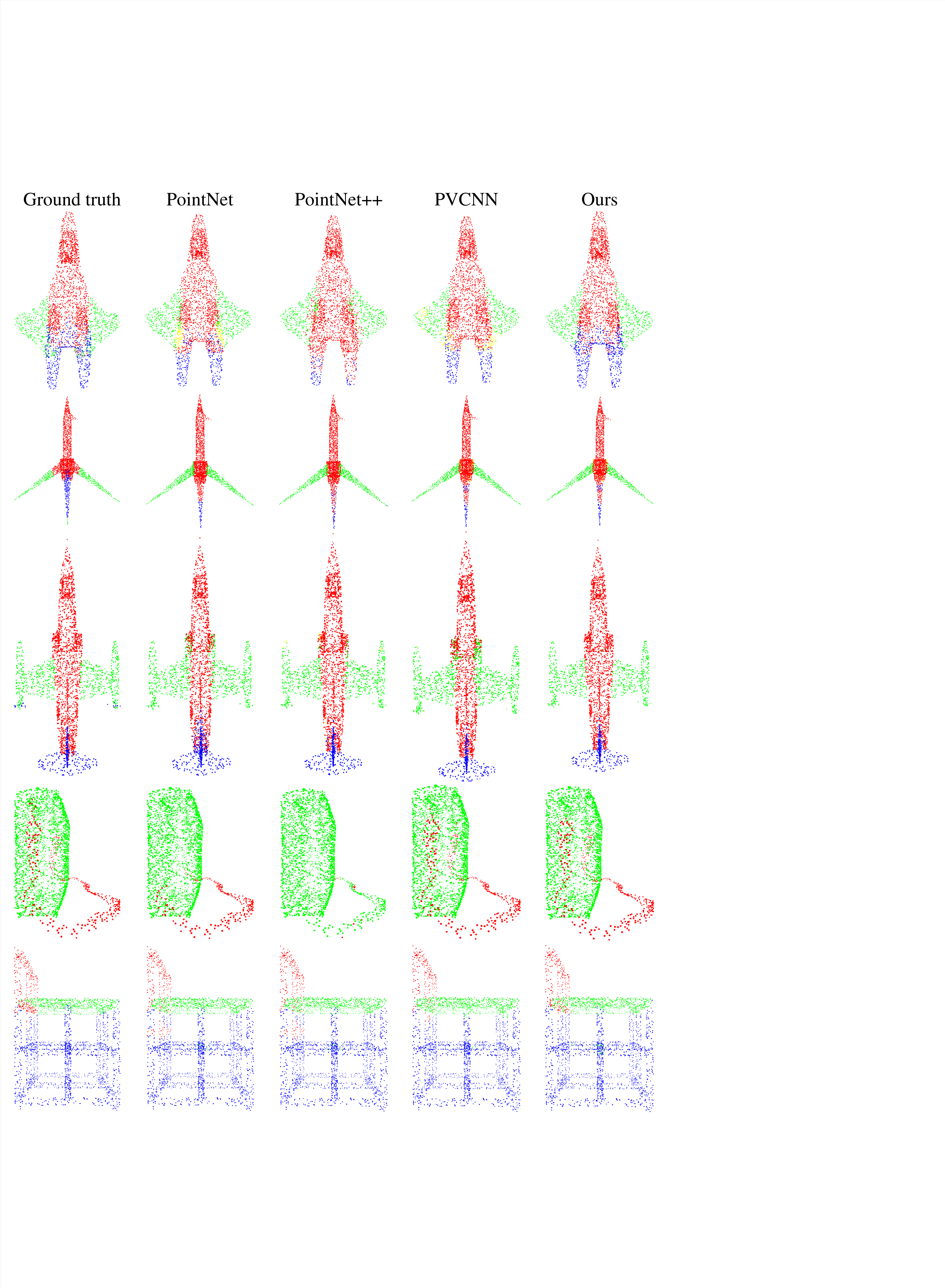}
	\end{center}
	\caption{Part segmentation results of PointNet~\cite{qi2017pointnet}, PointNet~\cite{qi2017pointnetplusplus} PVCNN~\cite{liu2019pvcnn} and our MPVCNN on Shapenet Part.}
	\label{fig:shapenet1}
\end{figure*}

\begin{figure*}[t]
	\begin{center}
		\includegraphics[width=0.8\linewidth]{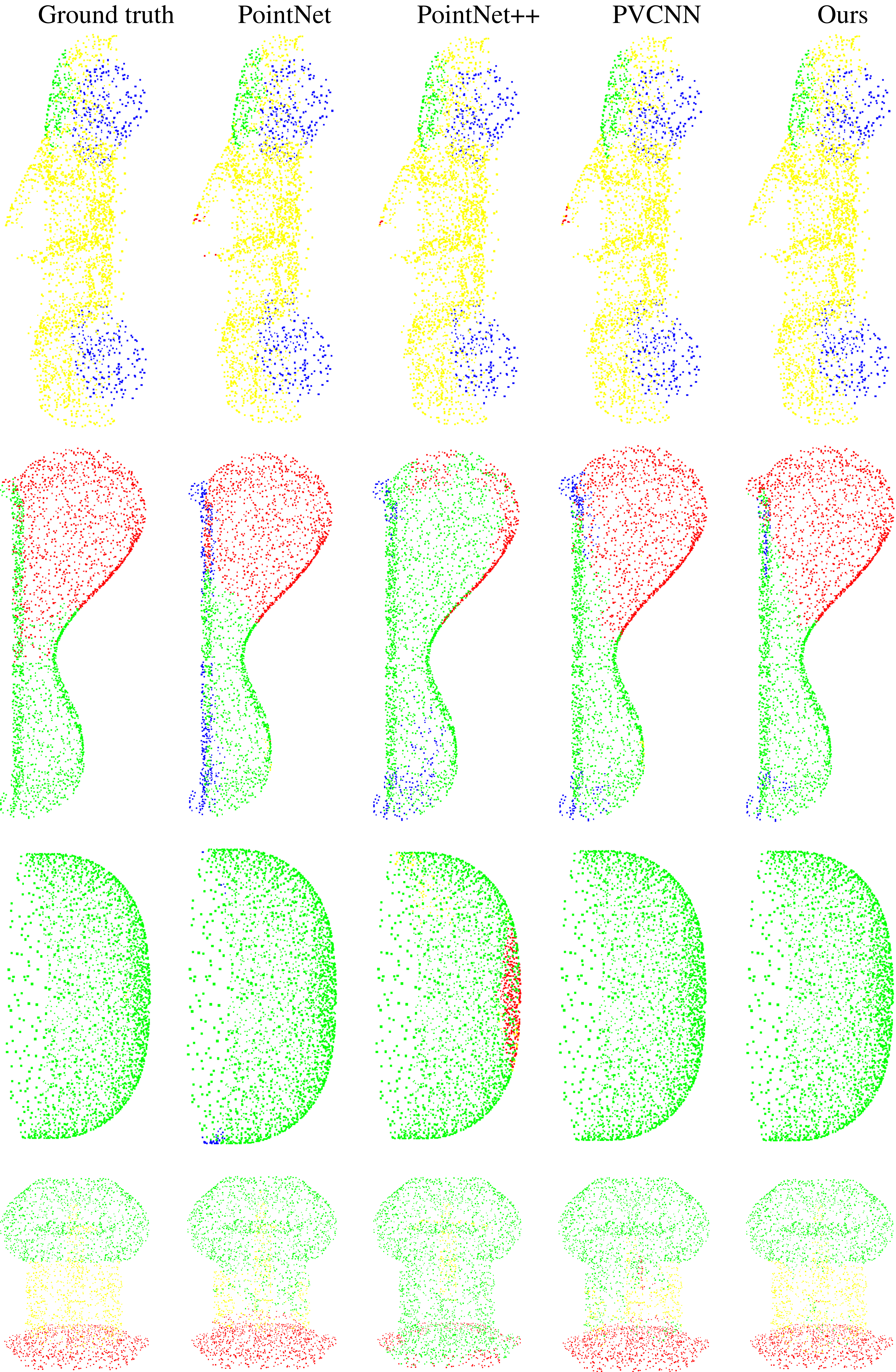}
	\end{center}
	\caption{Part segmentation results of PointNet~\cite{qi2017pointnet}, PointNet~\cite{qi2017pointnetplusplus} PVCNN~\cite{liu2019pvcnn} and our MPVCNN on Shapenet Part.}
	\label{fig:shapenet2}
\end{figure*}

\begin{figure*}[t]
	\begin{center}
		\includegraphics[width=0.85\linewidth]{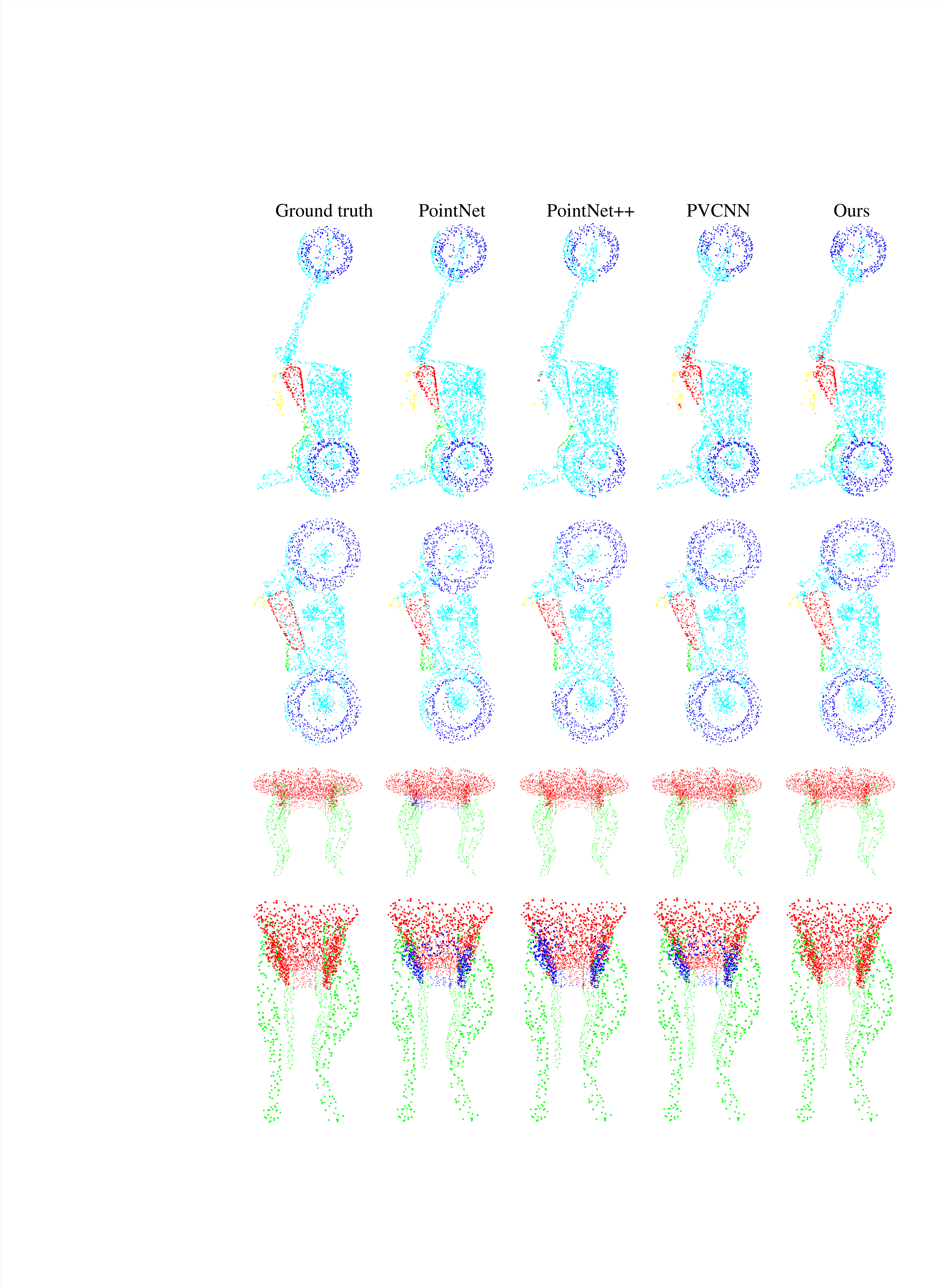}
	\end{center}
	\caption{Part segmentation results of PointNet~\cite{qi2017pointnet}, PointNet~\cite{qi2017pointnetplusplus} PVCNN~\cite{liu2019pvcnn} and our MPVCNN on Shapenet Part.}
	\label{fig:shapenet3}
\end{figure*}

\begin{figure*}[t]
	\begin{center}
		\includegraphics[width=1\linewidth]{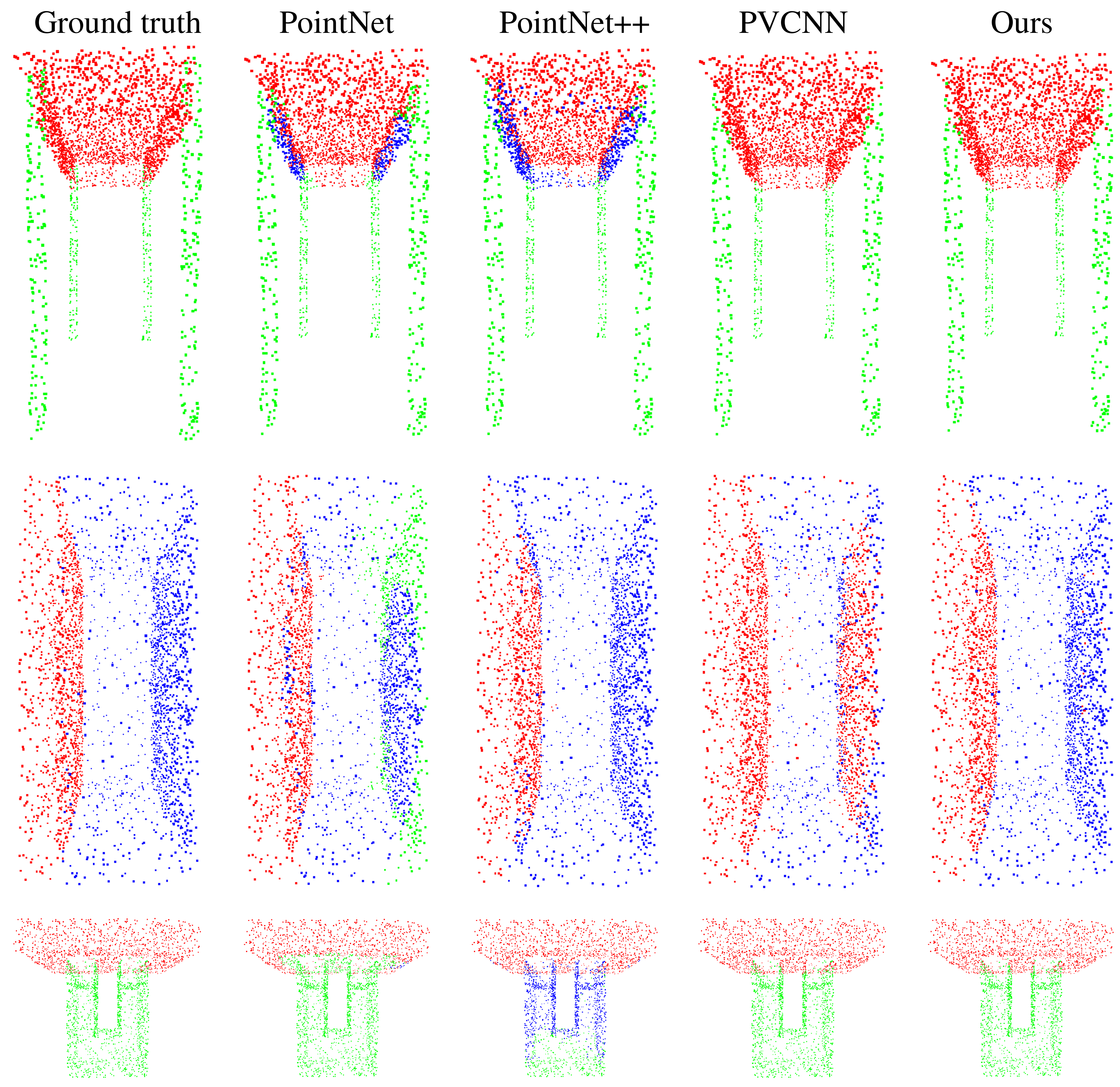}
	\end{center}
	\caption{Part segmentation results of PointNet~\cite{qi2017pointnet}, PointNet~\cite{qi2017pointnetplusplus} PVCNN~\cite{liu2019pvcnn} and our MPVCNN on Shapenet Part.}
	\label{fig:shapenet4}
\end{figure*}

\begin{figure*}[t]
	\begin{center}
		\includegraphics[width=0.81\linewidth]{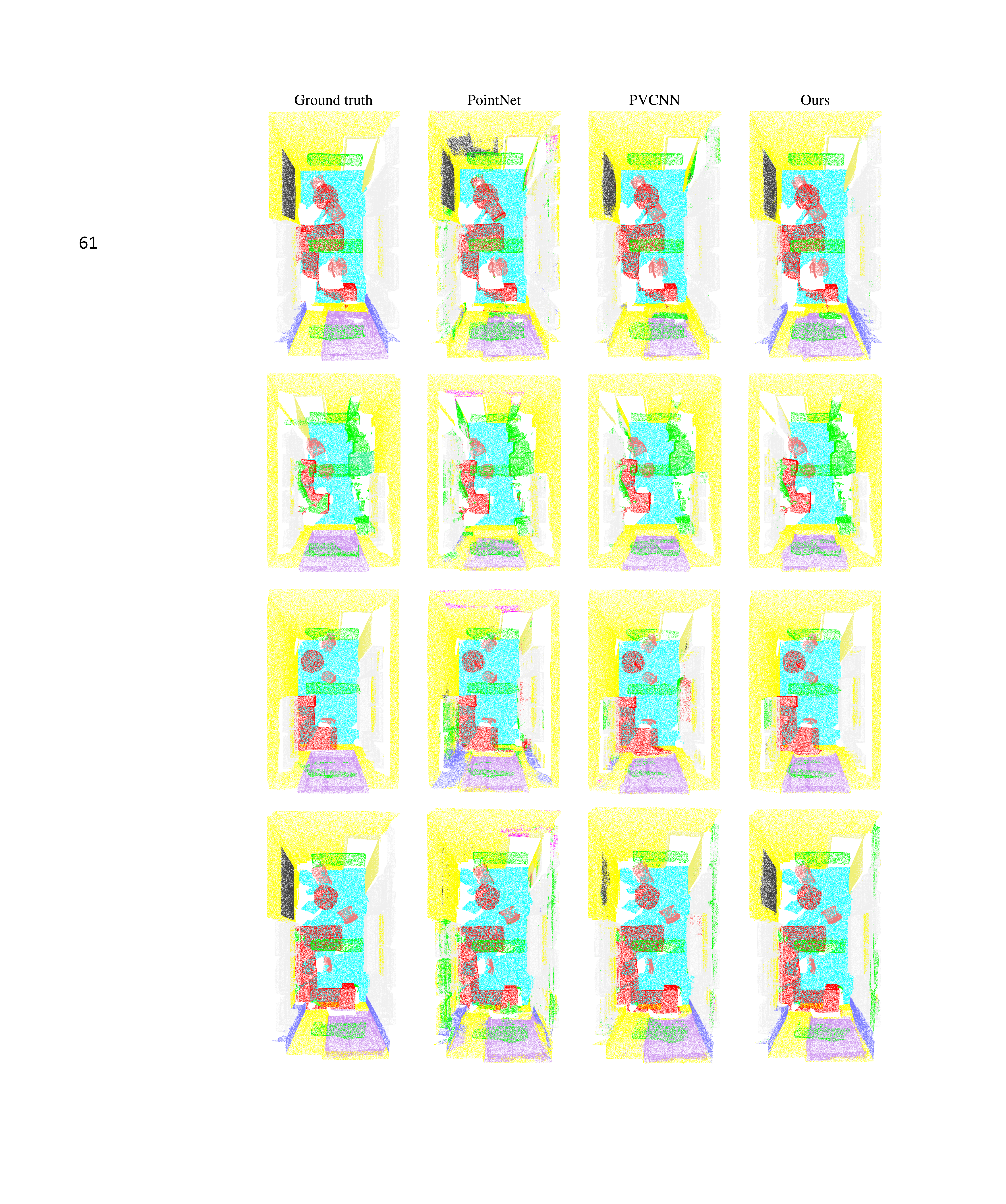}
	\end{center}
	\caption{Indoor scene segmentation results of PointNet~\cite{qi2017pointnet}, PVCNN~\cite{liu2019pvcnn} and our MPVCNN on S3DIS are5.}
	\label{fig:s3dis1}
\end{figure*}

\begin{figure*}[t]
	\begin{center}
		\includegraphics[width=0.93\linewidth]{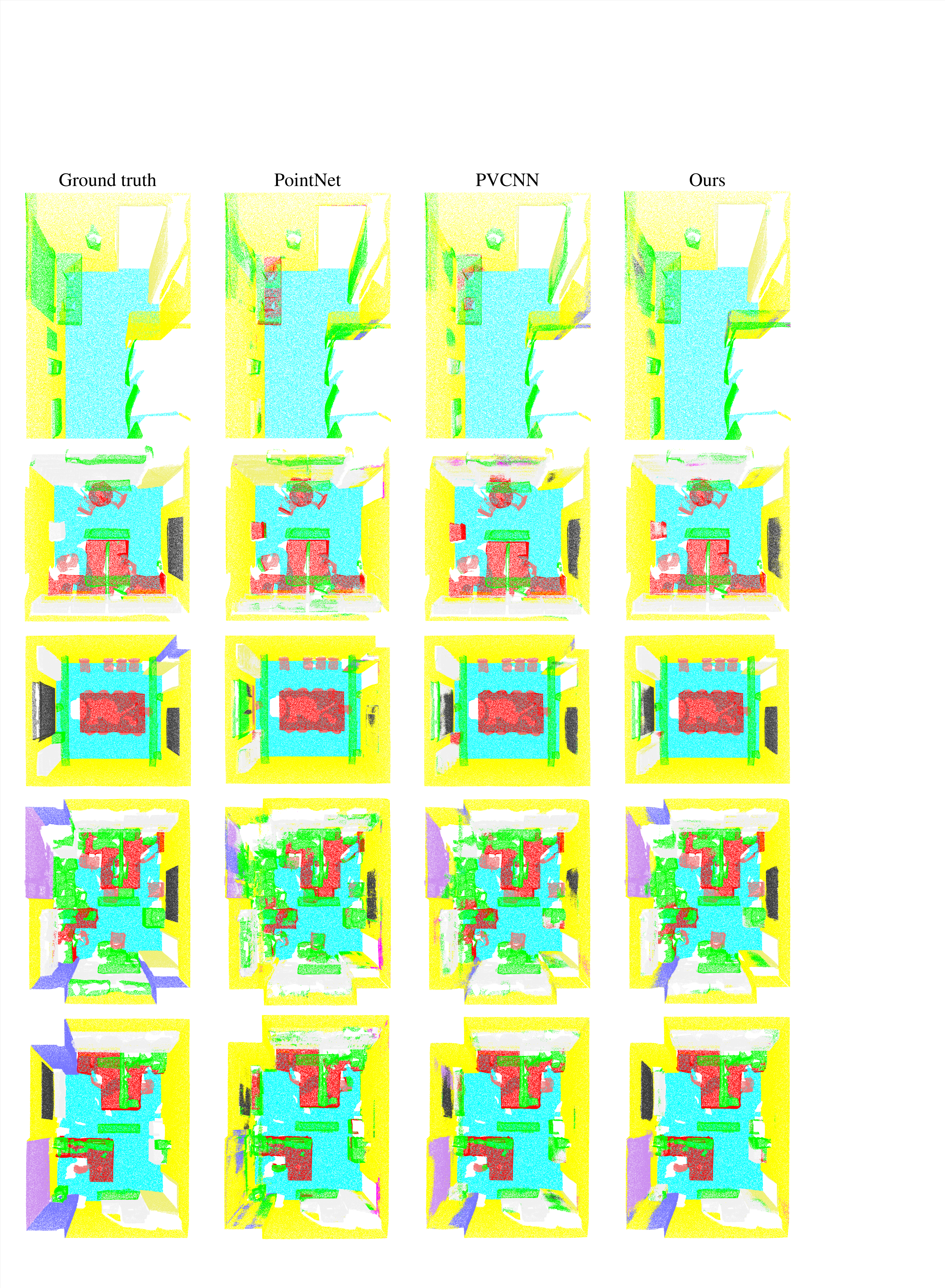}
	\end{center}
	\caption{Indoor scene segmentation results of PointNet~\cite{qi2017pointnet}, PVCNN~\cite{liu2019pvcnn} and our MPVCNN on S3DIS are5.}
	\label{fig:s3dis2}
\end{figure*}

\begin{figure*}[t]
	\begin{center}
		\includegraphics[width=0.78\linewidth]{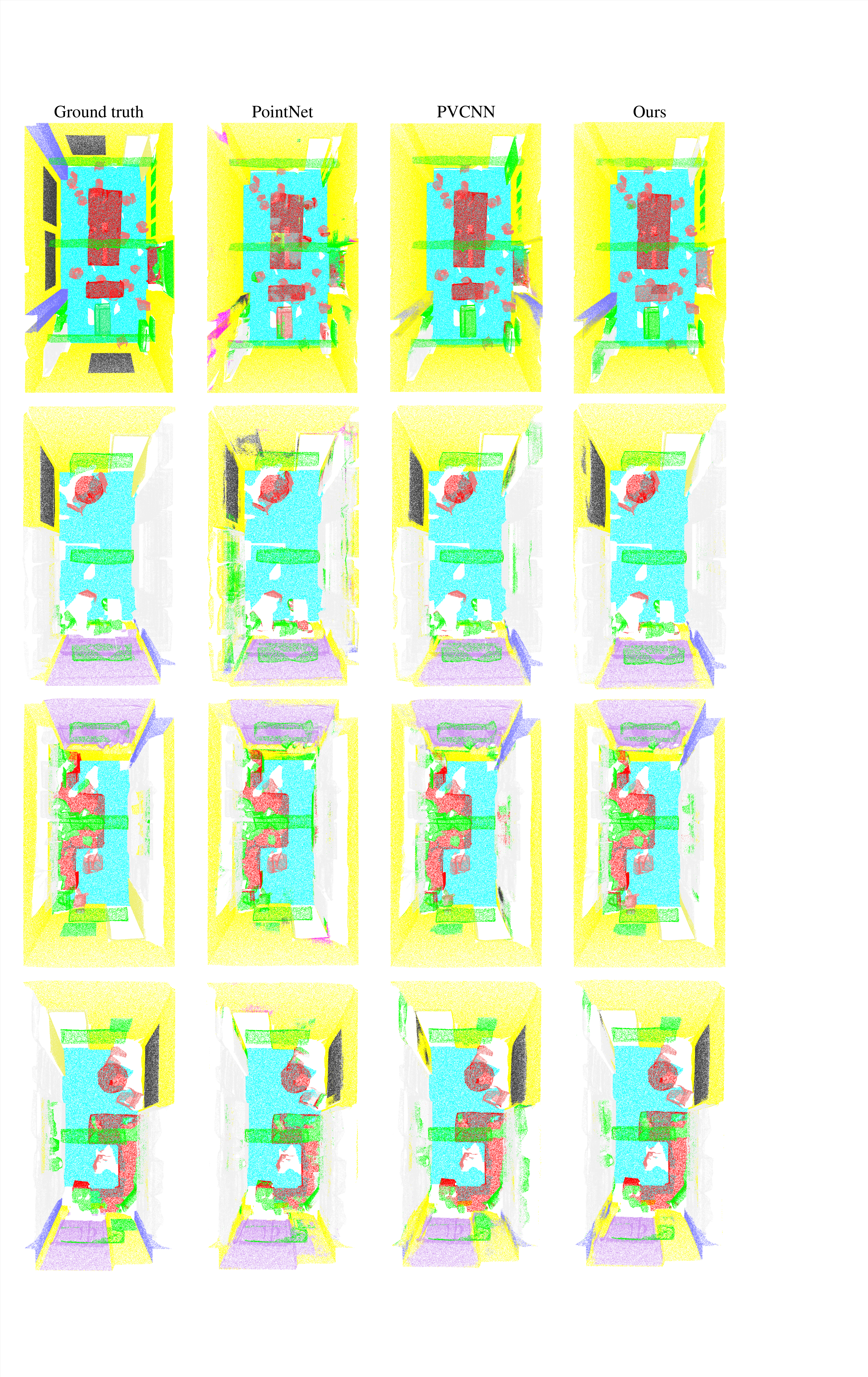}
	\end{center}
	\caption{Indoor scene segmentation results of PointNet~\cite{qi2017pointnet}, PVCNN~\cite{liu2019pvcnn} and our MPVCNN on S3DIS are5.}
	\label{fig:s3dis3}
\end{figure*}

\begin{figure*}[t]
	\begin{center}
		\includegraphics[width=0.77\linewidth]{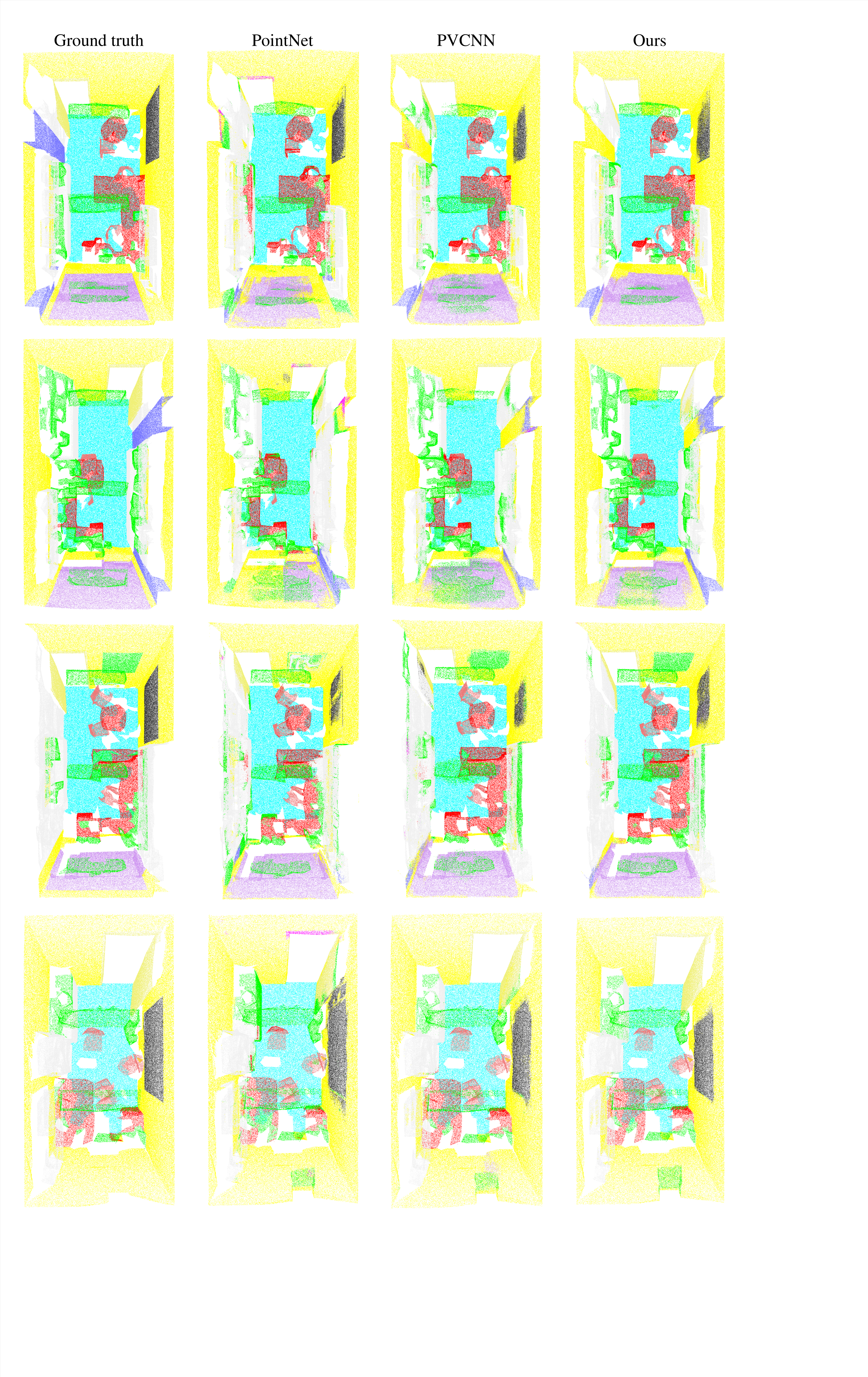}
	\end{center}
	\caption{Indoor scene segmentation results of PointNet~\cite{qi2017pointnet}, PVCNN~\cite{liu2019pvcnn} and our MPVCNN on S3DIS are5.}
	\label{fig:s3dis4}
\end{figure*}

\begin{figure*}[t]
	\begin{center}
		\includegraphics[width=0.82\linewidth]{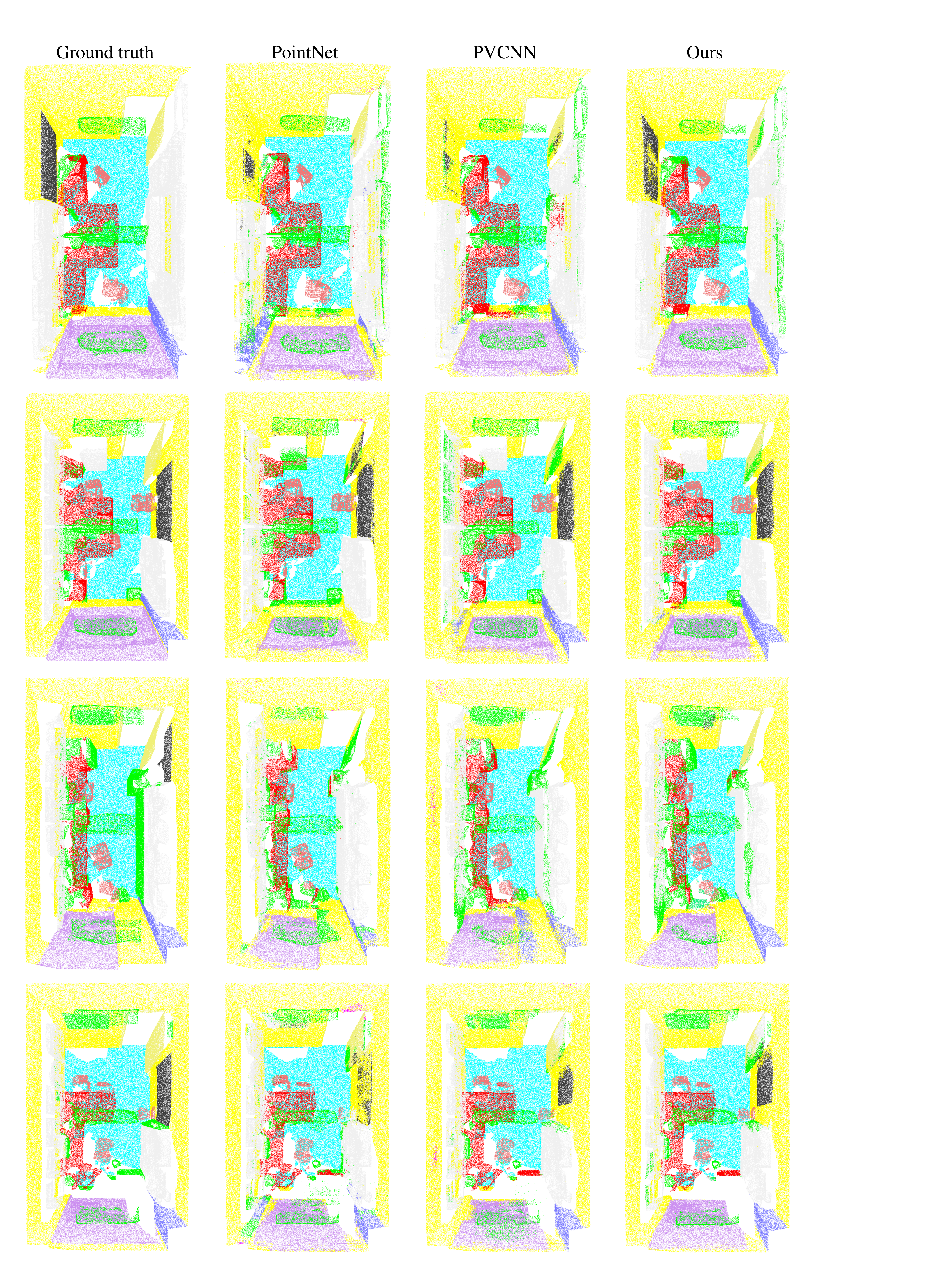}
	\end{center}
	\caption{Indoor scene segmentation results of PointNet~\cite{qi2017pointnet}, PVCNN~\cite{liu2019pvcnn} and our MPVCNN on S3DIS are5.}
	\label{fig:s3dis5}
\end{figure*}

In the supplementary materials, we show more comparable visualization results in part segmentation (Figure~\ref{fig:shapenet1}, Figure~\ref{fig:shapenet2}, Figure~\ref{fig:shapenet3} and Figure~\ref{fig:shapenet4}) and indoor scene segmentation (Figure~\ref{fig:s3dis1}, Figure~\ref{fig:s3dis2}, Figure~\ref{fig:s3dis3}, Figure~\ref{fig:s3dis4} and Figure~\ref{fig:s3dis5}) tasks with PointNet~\cite{qi2017pointnet}, PointNet++~\cite{qi2017pointnetplusplus}, PVCNN~\cite{liu2019pvcnn} and our MPVCNN. 
Besides, we will make our code and models publicly available.

\end{document}